\documentclass[10pt,twocolumn,letterpaper]{article}

\usepackage{cvpr}
\usepackage{times}
\usepackage{epsfig}
\usepackage{graphicx}
\usepackage{amsmath}
\usepackage{amssymb}
\usepackage{amsthm}
\usepackage{subfigure}
\usepackage{caption}
\usepackage{enumitem}
\usepackage{algorithm}
\usepackage{multirow}
\usepackage{balance}
\usepackage{algorithmic}
\usepackage{color, colortbl}
\newtheorem{definition}{Definition}
\algsetup{linenosize=\small}
\pdfoutput=1

\usepackage[pagebackref=false,breaklinks=true,letterpaper=true,colorlinks,bookmarks=false]{hyperref}

\cvprfinalcopy 

\def\cvprPaperID{****} 

\ifcvprfinal\fi
\definecolor{Gray}{gray}{0.9}

\newcommand{\SKIP}[1]{} 

\newcommand{\mbegin} {\left [ \begin{array}}

\newcommand{\mend}   {\end{array} \right ]}

\newcommand{\detbegin} {\left | \begin{array}}
\newcommand{\detend}   {\end{array} \right |}

\newcommand{\vbegin} {\left ( \begin{array}{c}}

\newcommand{\vend} {\end{array}\right )}

\def\squareforqed{\hbox{\rlap{$\sqcap$}$\sqcup$}}

\def\qed{\ifmmode\squareforqed\else{\unskip\nobreak\hfil
	\penalty50\hskip1em\null\nobreak\hfil\squareforqed
	\parfillskip=0pt\finalhyphendemerits=0\endgraf}\fi}

\newcommand{\showeqnlabel}{
	\hbox to 0pt{\quad\quad\relax\fbox{\scriptsize\rm\eqnlblx}%
	\gdef\eqnlblx{xxxx}}} \newcommand{\eqnlblx}{}

\def\@eqnnum{\rm (\theequation)\showeqnlabel}

\newcommand{\nofig}[1]{\centerline{\bf Figure here}}

\def\mat#1{\mathchoice{\mbox{\bf$\displaystyle\tt#1$}}
	{\mbox{\bf$\textstyle\tt#1$}}
	{\mbox{\bf$\scriptstyle\tt#1$}}
	{\mbox{\bf$\scriptscriptstyle\tt#1$}}}
\def\m#1{\protect\mat #1}

\DeclareRobustCommand{\rchi}{{\mathpalette\irchi\relax}}
\newcommand{\irchi}[2]{\raisebox{\depth}{$#1\chi$}} 
\begin{document}

\title{Jumping Manifolds: Geometry Aware Dense Non-Rigid Structure from Motion}
\author{Suryansh Kumar\\
Australian National University, CECS, Canberra, ACT, 2601\\
{\tt\small suryansh.kumar@anu.edu.au}
}

\maketitle

\begin{abstract}
Given dense image feature correspondences of a non-rigidly moving object across multiple frames, this paper proposes an algorithm to estimate its 3D shape for each frame.  To solve this problem accurately, the recent state-of-the-art algorithm reduces this task to set of local linear subspace reconstruction and clustering problem using Grassmann manifold representation \cite{kumar2018scalable}. Unfortunately, their method missed on some of the critical issues associated with the modeling of surface deformations, for e.g., the dependence of a local surface deformation on its neighbors. Furthermore, their representation to group high dimensional data points inevitably introduce the drawbacks of categorizing samples on the high-dimensional Grassmann manifold \cite{huang2015projection, harandi2014manifold}. Hence, to deal with such limitations with \cite{kumar2018scalable}, we propose an algorithm that jointly exploits the benefit of high-dimensional Grassmann manifold to perform reconstruction, and its equivalent lower-dimensional representation to infer suitable clusters. To accomplish this, we project each Grassmannians onto a lower-dimensional Grassmann manifold which preserves and respects the deformation of the structure w.r.t its neighbors. These Grassmann points in the lower-dimension then act as a representative for the selection of high-dimensional Grassmann samples to perform each local reconstruction. In practice, our algorithm provides a geometrically efficient way to solve dense NRSfM by switching between manifolds based on its benefit and usage. Experimental results show that the proposed algorithm is very effective in handling noise with reconstruction accuracy as good as or better than the competing methods.
\end{abstract}


\section{Introduction}
\emph{Non-rigid Structure-from-Motion (NRSfM)}, a problem where the task is to recover the three-dimensional structure of a deforming object from a set of image feature correspondences across frames. Any solution to this problem depends on the proper modeling of \emph{structure} $\in \mathcal{M}$ and an efficient estimation of \emph{motion} $\in \mathbb{SE}(3)$, where $\mathcal{M}$ denotes some structure manifold and $\mathbb{SE}(3)$ denotes special Euclidean group which is a differentiable manifold \cite{fecko2006differential}. Though, after Bregler \etal factorization framework to NRSfM \cite{tomasi1992shape}, motion estimations are mostly relaxed to rotation estimation $\in \mathbb{SO}(3)$. 
Even after such relaxation, the problem still remains unsolved for any arbitrary motion. The main difficulty in NRSfM comes from the fact that both the camera and the object are moving and, along with it the object themselves are deforming, hence, it becomes difficult to distinguish camera motion from object motion using only image data. Despite such difficulties, many efficient and reliable solutions based on the priors are proposed to solve NRSfM. A reliable solution to this problem is important as it covers a wide range of applications from medical industry to the entertainment industry and many more.
\begin{figure}[t]
\begin{center}
\includegraphics[width=0.7\linewidth, height= 0.55\linewidth]{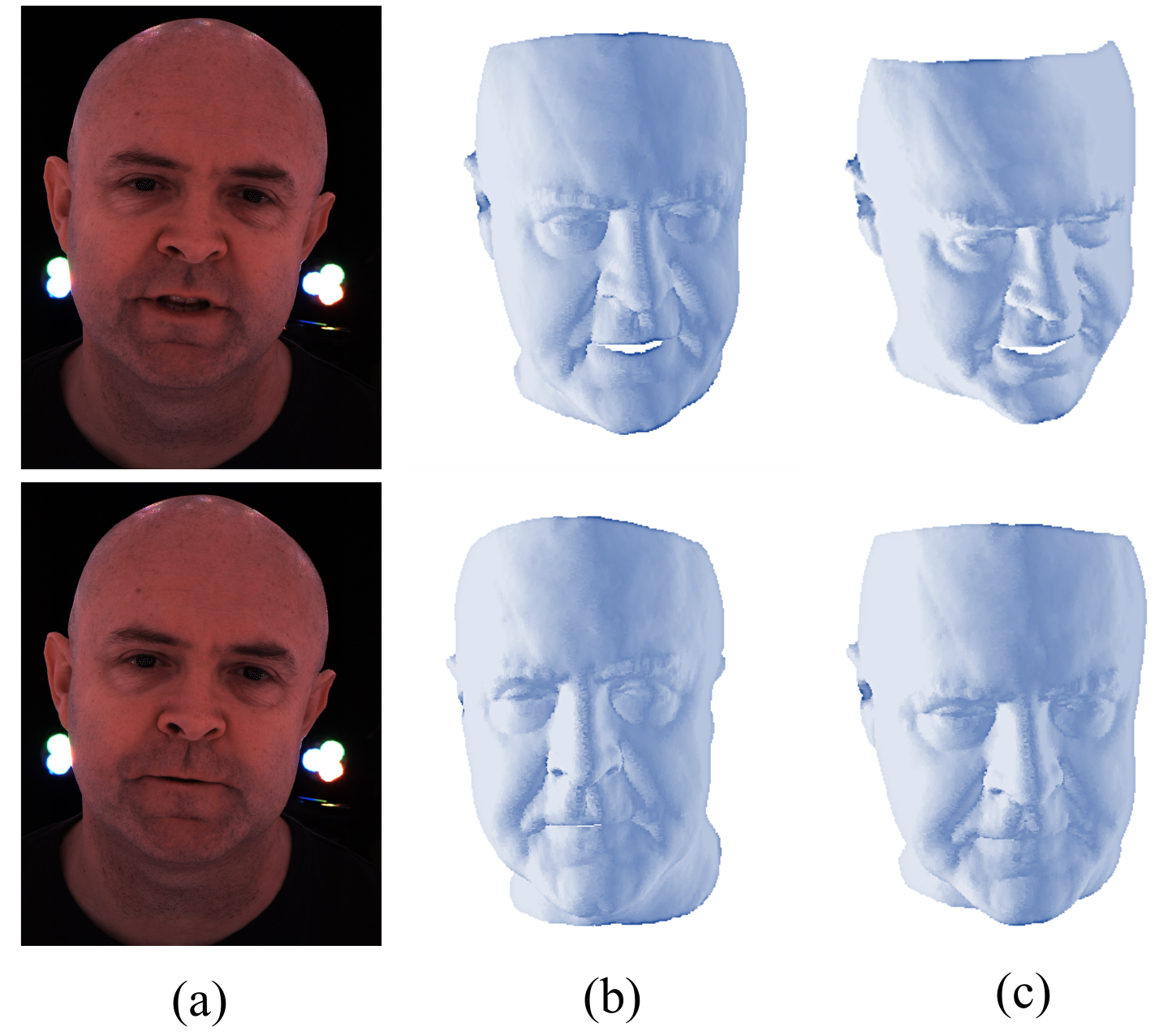}
\caption{\small{Dense 3D reconstruction of facial expression using our algorithm. The result show the 3D reconstruction of 73,765 points of a complex non-rigidly deforming surface. These results can be useful for real world applications such as 3D modeling, virtual reality etc. The example sequence is taken from Actor dataset \cite{beeler2011high}. } \label{fig:firstPage}}
\end{center}
\end{figure}

To solve NRSfM, the algorithms proposed in the past can broadly be divided into two major classes 1) \emph{sparse} NRSfM and 2) \emph{dense} NRSfM. This classification is based on the number of feature points that the algorithm can efficiently process to model the deformation of the object. Although many reliable solution to this problem exists for sparse NRSfM \cite{dai2014simple, kumar2017spatio, kumar2016multi,  akhter2011trajectory, taylor2010non,paladini2012optimal,lee2013procrustean,gotardo2011non,hamsici2012learning}, very few work have been done towards solving the dense NRSfM reliably and efficiently \cite{garg2013dense, dai2017dense, kumar2018scalable, ansari2017scalable}. Also, the existing solutions to dense NRSfM are computationally expensive and are mostly constrained to analyze the global deformation of the non-rigid shape \cite{garg2013dense,dai2017dense}. 
The basis for this gradual progress in dense NRSfM is perhaps due to its dependence on per pixel reliable correspondences across frames, and the absence of a resilient structure modeling framework to capture the local non-linearities. One may argue on the efficient motion estimation, however, from image correspondences, we can only estimate relative motion and reliable algorithms with solid theory exists to perform this task well \cite{dai2014simple,lee2013procrustean}. Also, with the recent progress in deep learning algorithms, per pixel correspondences can be achieved with a remarkable accuracy \cite{sun2018pwc}, which leaves structure modeling as a potential gray area in dense NRSfM to focus.

\begin{figure}[t]
\begin{center}
\includegraphics[width=0.70\linewidth]{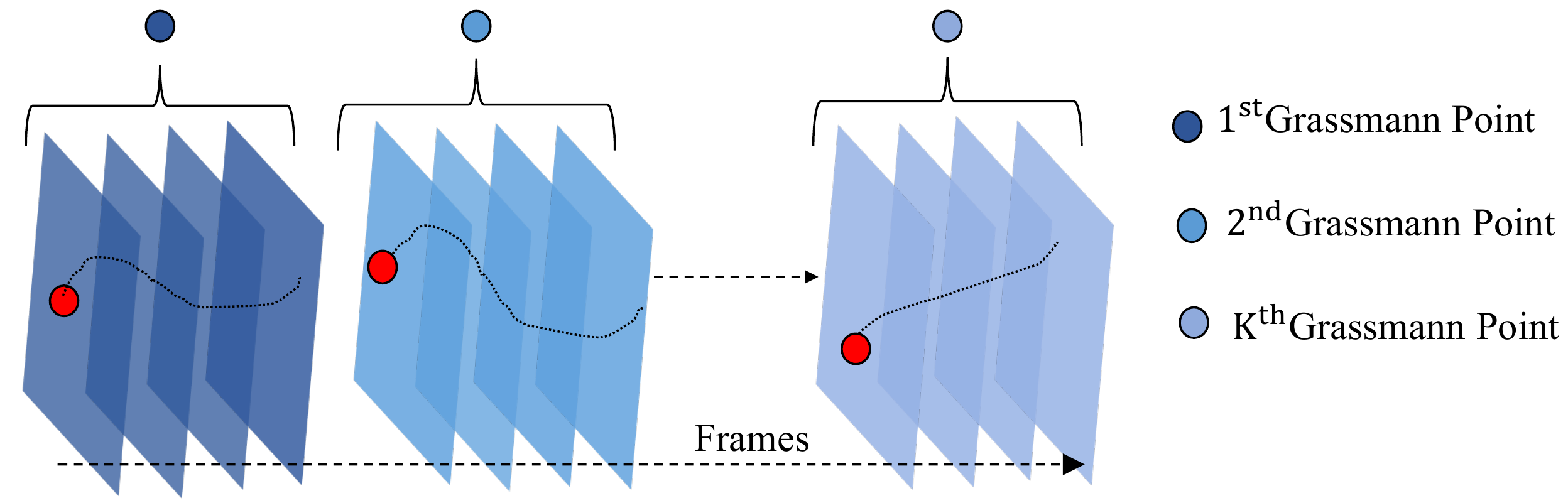}
\caption{\small{Temporal representation using Grassmannians in the shape space introduces discontinuity in the overall trajectory of the feature point. Also, to define neighboring subspace dependency graph in the time domain seems very challenging keeping in mind that the activity/expression may repeat.  Red circle shows the feature point with its trajectory over frames (Black). } \label{fig:temporaldis}}
\end{center}
\end{figure}

Very recently, Kumar \etal \cite{kumar2018scalable} has exploited the Grassmann manifold to model non-rigid surfaces in dense NRSfM. The key insight in their work is; even though the overall complexity of the deforming shape is high, each local deformation may be less complex \cite{crane2013conformal, crane2015discrete, crane2013digital, Crane:2017:GID}. Using this idea, they proposed a union of local linear subspace approach to solve dense NRSfM problem. Nevertheless, their work overlooked on some of the intrinsic issues associated with the modeling of non-rigidly deforming surface. \emph{Firstly}, their method represents each local linear subspace independently via a high-dimensional Grassmannian representation. Now, such representation may help reconstruct complex 3D deformation but can lead to wrong clustering, and it's very important in joint reconstruction and clustering framework to have suitable clustering of subspaces, else reconstruction may suffer. \emph{Secondly}, their approach to represent local non-linear deformation completely ignored the neighboring surfaces, which may result in an inefficient representation of the Grassmannians in the trajectory space. \emph{Thirdly}, the representation of Grassmannians in the shape space adopted by \cite{kumar2018scalable} results in \emph{irredeemable} discontinuity of the trajectories (see Fig.(\ref{fig:temporaldis})). Hence, temporal representation of the set of shapes using Grassmannians seems not an extremely beneficial choice for modeling dense NRSfM on Grassmannian manifold\footnote{Purpose behind NRSfM is not the same as activity/action recognition. See supplementary material for a detail discussion on this.}. \emph{Lastly}, although the dense NRSfM algorithm proposed in \cite{kumar2018scalable} works better and faster than the previous methods, it depends on several manual parameters which are inadmissible for practical applications.

This paper introduces an algorithm that overcomes the aforementioned limitations with Kumar \etal method \cite{kumar2018scalable}. The main point we are trying to make is that; reconstruction and grouping of subspace on the same high dimensional Grassmann manifold seem an unreasonable choice. Even recent research in the Riemannian geometry has shown that the low-dimensional representation of the corresponding high dimensional Grassmann manifold is more favorable for grouping Grassmannians \cite{huang2015projection,harandi2014manifold}. So, we formulate dense NRSfM in a way that it takes advantage of both high and low dimensional representation of Grassmannians \ie, perform reconstruction in the original high-dimension and cluster subspace in its lower-dimension representation.

We devise an unsupervised approach to efficiently represent the high-dimensional non-rigid surface on a lower dimensional Grassmann manifold. These low-dimensional Grassmannians are represented in such a way that it preserves the local structure of the surface deformation in accordance with its neighboring surfaces when projected. 
Now, these low-dimensional Grassmannians serves as a potential representative for its high-dimensional Grassmannians for suitable grouping, which subsequently help improve the reconstruction and representation of the Grassmannians on the high-dimensional Grassmann manifold, hence, the term \emph{{\bf{J}}\textcolor{red}{u}mping {\bf{M}}anif\textcolor{red}{o}lds} ({\bf{M}}\textcolor{red}{o}{\bf{J}}\textcolor{red}{u}). Further, we drop the temporal grouping of shapes using Grassmannians to discourage the discontinuity of trajectories (see Fig.(\ref{fig:temporaldis})). 

In essence, our work is inspired by \cite{kumar2018scalable} and is oriented towards settling its important limitations. Moreover, in contrast to \cite{kumar2018scalable}, we capture the notion of dependent local subspace in a union of subspace algorithm \cite{larsson2017compact} via Grassmannian modeling. The algorithm we proposed is an attempt to supply a more efficient, reliable and practical solution to this problem. Our formulation gives an efficient framework for modeling dense NRSfM on the Grassmann manifold than \cite{kumar2018scalable}. Experimental results show that our method is as accurate as other algorithms and is numerically more efficient in handling noise. The main contributions of our work are as follows:
\begin{itemize}[noitemsep]
\item An efficient framework for modeling non-rigidly deforming surface that exploits the advantage of Grassmann manifold representation of different dimensions based on its geometry.
\item A formulation that encapsulates the local non-linearity of the deforming surface w.r.t its neighbors to enable the proper inference and representation of local linear subspaces. 
\item An iterative solution to the proposed cost function based on ADMM \cite{boyd2011distributed}, which is simple to implement and provide results as good as the best available methods. Additionally, it helps improve the 3D reconstruction substantially, in the case of noisy trajectories.
\end{itemize}

\section{Related Work}
A working solution to NRSfM was first introduced in the seminal work by Bregler \etal~\cite{bregler2000recovering} which is an extension to Tomsai-Kannade rigid factorization method\cite{tomasi1992shape}. Although the problem still remains unsolved for arbitrary deformations, many profound works have been done to achieve a reliable solution to this problem under some or the other prior assumptions about the object \cite{dai2014simple, akhter2011trajectory, park20103d, lee2013procrustean, zhu2014complex, kumar2017spatio, kumar2016multi, garg2013dense}. Since the literature on this topic is very extensive, we review the works that are of close relevance to the {dense} NRSfM methods under classical setting\footnote{By \emph{classical setting}, we mean without using RGB-D or 3D template.}.


Earlier attempts to solve dense NRSfM used piecewise reconstruction of the shape parts which were further processed via a stitching step to get a global 3D shape \cite{collins2010locally,russell2012dense}. To our knowledge, Garg \etal variational approach \cite{garg2013variational} was the first to propose and demonstrate per pixel dense NRSfM algorithm without any 3D template prior. This method introduced a discrete total variational constraint with trace norm constraint on the global shape, which resulted in a biconvex optimization problem. Despite the algorithm outstanding results, it's computationally very expensive and needs a GPU to provide the solution.

In contrast, Dai \etal extended his simple prior free approach to solve dense NRSfM problem~\cite{dai2014simple, dai2017dense}. The algorithm proposed a spatial-temporal formulation to solve the problem. The author revisits the temporal smoothness term from \cite{dai2014simple} and integrate it with a spatial smoothness term using the Laplacian of the non-rigid shape. The resultant optimization leads to a series of least squares to be minimized which makes it extremely slow to process. Recently, Kumar \etal modeled this problem on the Grassmann manifold \cite{kumar2018scalable}. Their work extended the spatiotemporal multi-body framework to solve dense NRSfM \cite{kumar2017spatio}. The algorithm demonstrated that such an approach is more efficient, faster and accurate than all the other recent approaches to solve dense NRSfM task \cite{garg2013variational, dai2017dense,ansari2017scalable}.  

Consecutive frame-based approach has recently shown some promising results to solve dense 3D reconstruction of a general dynamic scene including non-rigid object \cite{kumar2017monocular,ranftl2016dense}. Nevertheless, motion segmentation, triangulation, as rigid as possible constraint and scale consistency quite often breaks down for the deforming object over frames. Therefore, dense NRSfM becomes extremely challenging for such algorithms. Not long ago, Gallardo \etal combined shading, motion and generic physical deformation to model dense NRSfM \cite{gallardo2017dense}.


\section{Preliminaries}
In this paper, $\| .\|_{\m F}$, $\| .\|_{*}$ denotes the Frobenius norm and nuclear norm respectively. $\| .\|_{\mathcal{G}}$ represent the notion of norm on the Grassmann manifold. Single angle bracket $<.,.>$ denotes the Euclidean inner product. For ease of understanding and completeness, in this section, we briefly review few important definitions related to the Grassmann manifold. Firstly, a manifold is a topological space that is locally similar to the Euclidean space. Out of several manifolds, the Grassmann manifold is a topologically rich non-linear manifold, each point of which represent the set of all right invariant subspace of the Euclidean space \cite{dollar2007non,absil2009optimization,kumar2018scalable}.
\begin{definition}\label{def:1}
   The Grassmann manifold, denoted by $\mathcal{G}(\m p, \m d)$, consists of all the linear $\m p$-dimensional subspace embedded in a `$\m d$' dimensional Euclidean space $\mathbb{R}^{\m d}$ such that $0\leq \m p \leq \m d$ [Absil et al., 2009] \cite{absil2009optimization}.
\end{definition}
A point `$\Phi$' on the Grassmann manifold can be represented by $ \mathbb{R}^{\m d \times \m p}$ matrix whose columns are  composed of orthonormal basis. The space of such matrices with orthonormal columns is a Riemannian manifold such that $\Phi^{\m T} \Phi = \mathbf{I}_{\m p}$, where $\mathbf{I}_{\m p}$ is a $\m p \times \m p$ identity matrix. 

\begin{definition}\label{def:2}
	Grassmann manifold can be embedded into the space of symmetric matrices via mapping $\Pi: \mathcal{G}(\m p, \m d) \mapsto \textrm{Sym}(\m d), \Pi(\Phi) = \Phi\Phi^{\m T}$, where $\Phi$ is a Grassmann point \cite{hamm2008grassmann, harandi2013dictionary}.
   Given two Grassmann points $\Phi_1$ and $\Phi_2$, then the distance between them can be measured using the projection metric $d_g^2(\Phi_1,  \Phi_2) = 0.5\|\Pi(\Phi_1) - \Pi(\Phi_2)\|_{\m F}^{2}$ \cite{hamm2008grassmann}.
\end{definition}

In the past, these two properties of Grassmann manifold has been used in many computer vision applications \cite{hamm2008grassmann,cetingul2009intrinsic,kumar2018scalable}. Second definition is very important as it allows to measure the distance on the Grassmann manifold, hence, $(\mathcal{G}, d_g$) forms a metric space. We used these properties in the construction of our formulation. For comprehensive details on this topic readers may refer to \cite{hamm2008grassmann}.

\section{Problem Formulation}
Let `$\m P$' be the total number of feature points tracked across `$\m F$' frames. Concatenating these 2D coordinates of each feature points for all frames across the columns of a matrix gives `$\m W$' $\in \mathbb{R}^{2\m F \times \m P}$ matrix. This matrix is popularly known as \emph{measurement matrix} \cite{tomasi1992shape}. Our goal is, given the image measurement matrix, estimate the camera motion and 3D coordinates of every 2D feature points across all frames. 

We start our formulation with the classical representation to NRSfM \ie $\m W = \m R \m S$, where, $\m R \in \mathbb{R}^{2\m F \times 3\m F}$ is a block diagonal rotation matrix with each block as a $2 \times 3$ orthographic rotation matrix, and $\m S \in \mathbb{R}^{3\m F \times \m P}$ as the 3D structure matrix. With such a representation, the entire problem simplifies to the estimation of correct rotation matrix `$\m R$' and structure matrix `$\m S$' such that the above relation holds. Following the assumption of the previous work \cite{kumar2018scalable}, we estimate the rotation using Intersection method \cite{dai2014simple}. As a result, the task reduces to
composing of an efficient algorithm that correctly models the surface deformations and provide better reconstruction results. Recent algorithms in NRSfM have demonstrated that clustering benefits reconstruction and vice-versa, however, the existing framework to employ this idea is not scalable to millions of points. To establish this idea for dense NRSfM, Kumar \etal \cite{kumar2018scalable} used LRR on Grassmannian manifold. Using the similar notions, we model dense deforming surface using Grassmannian representation to provide more reliable and accurate solution. 

In the following subsection, we first introduce the Grassmannian representation of the surface and how to project these Grassmannians onto the lower dimension Grassmann manifold by preserving the neighboring information. In the later subsection, we use these representations to formulate the overall cost function for solving dense NRSfM problem.


\subsection{Grassmannian representation}\label{ss:gr_rep}
\begin{figure}[t]
\begin{center}
\includegraphics[width=0.80\linewidth]{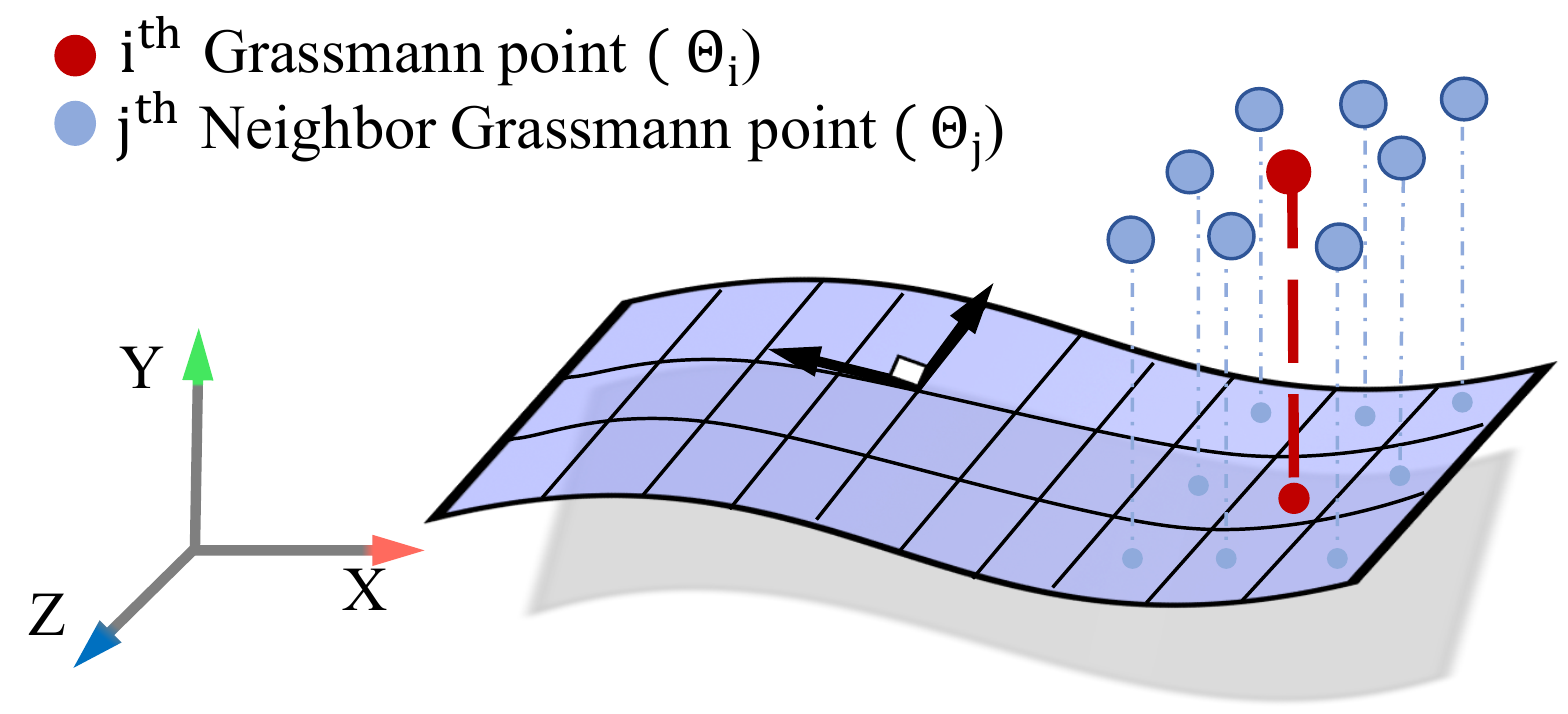}
\caption{\small{In contrast to \cite{kumar2018scalable}, our modeling of surface using Grassmannians considers the similarity between the neighboring Grassmannians while representing it in the lower dimension. Based on the assumption that spatially neighboring surface tend to span similar subspace, defining neighboring subspace dependency graph is easy and most of the real-world examples follows such assumption. However, building such graph in shape space can be tricky.} \label{fig:grassmannmodel1}}
\end{center}
\end{figure}

\begin{figure}[t]
\begin{center}
\includegraphics[width=0.90\linewidth]{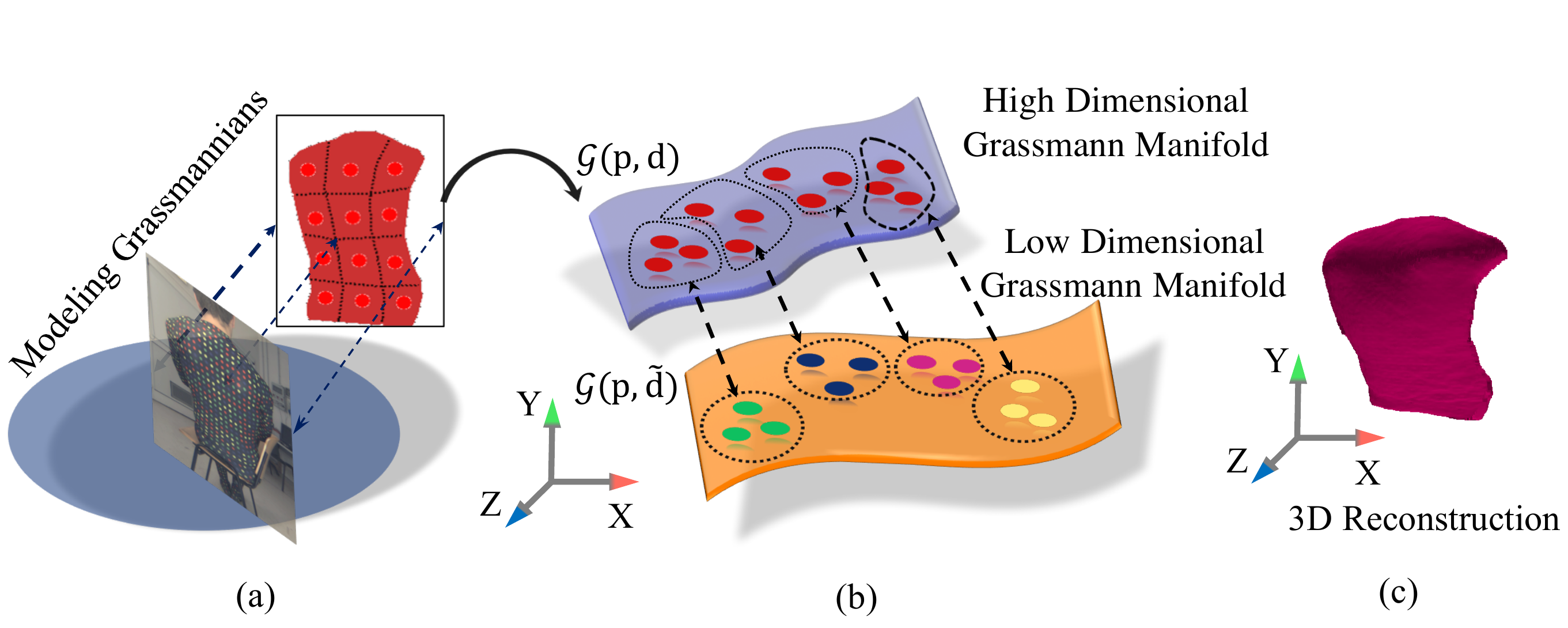}
\caption{\small{Conceptual illustration of our modeling (a) Modeling of 3D trajectories to Grassmann points (b) The two grassmann manifold and mapping of the points between them to infer better cluster index that leads to better reconstruction (c) The 3D reconstruction of the non-rigid deforming object.} \label{fig:grassmannmodel2}}
\end{center}
\end{figure}

Let `$\Phi_{\m i}$' $\in \mathcal{G}(\m p, {\m d})$ be a Grassmann point representing the $\m i^{\textrm{th}}$ local linear subspace spanned by $\m i^{\textrm{th}}$ set of columns of `$\m S$'. Using this notion, we decompose the entire trajectories of the structure into a set of `$\m K$' Grassmannians $\xi = \{\Phi_1, \Phi_2, \Phi_3, ...., \Phi_{\m K} \}$. Now, such a representation treats each subspace independently and therefore, its low-dimensional linear representation may not be suitable to capture the surface dependent non-linearity. To properly represent Grassmannian which respects the neighboring non-linearity in low-dimension, we introduce a different strategy to model non-rigid surface in low-dimension. For now, let $\Delta \in \mathbb{R}^{{\m d} \times \tilde{\m d} }$ be a matrix that maps `$\Phi_{\m i}$' $\in \mathcal{G}(\m p, {\m d})$ to  `$\phi_{\m i}$' $\in \mathcal{G}(\m p, \tilde{\m d})$ such that $ \tilde{\m d} < {\m d}$. Mathematically, 
\begin{equation}
\phi_{\m i} = \Delta^{\m T} \Phi_{\m i}
\end{equation}
Its quite easy to examine that $\phi_i$ is not a orthogonal matrix and, therefore, does not qualifies as a potential point on a Grassmann manifold. However, by performing a orthogonal-triangular (QR) decomposition of $\phi_i$, we estimate the new representative of $\phi_i$ on the Grassmann manifold of `$\tilde{\m d}$' dimension.
\begin{equation}\label{eq:2}
\Theta_{\m i} \m U_{\m i} = \mathbf{qr}(\phi_{\m i}) = \Delta^{\m T} \Phi_{\m i} 
\end{equation}
Here, $\mathbf{qr(.)}$ is a function that returns the QR decomposition of the matrix. The $\Theta_{\m i} \in \mathbb{R}^{\tilde{\m d} \times \m p}$ is an orthogonal matrix and $\m U_{\m i} \in \mathbb{R}^{\m p \times \m p} $ is the upper triangular matrix\footnote{Note: The value of $\tilde{\m d} \geq \m p$, Use $[\Theta_{\m i}, \m U_{\m i}]=\textbf{qr}(\phi_{\m i}, 0)$ in MATLAB to get a square $\m U_{\m i}$ matrix ($\m U_{\m i} \in \mathbb{R}^{\m p \times \m p}$)}. Using Eq.(\ref{eq:2}), we represent the equivalence of $\Phi_{\m i}$ in low dimension as

\begin{equation}\label{eq:3}
\begin{aligned}
& \displaystyle \Theta_{\m i} = \Delta^{\m T}(\Phi_{\m i} \m U_{\m i}^{-1})\\
& \displaystyle \Theta_{\m i} = \Delta^{\m T}\Omega_{\m i}
\end{aligned}
\end{equation}
where, $\Omega_{\m i} = \Phi_{\m i} \m U_{\m i}^{-1} \in \mathbb{R}^{{\m d} \times \m p}$. The key-point to note is that both $\Theta_{\m i}$ and $\phi_{\m i}$ has the same column space. In principle such a representation is useful however, it does not serve the purpose of preserving the non-linearity w.r.t its neighbors. In order to encapsulate the local dependencies (see Fig.(\ref{fig:grassmannmodel1}), Fig.(\ref{fig:grassmannmodel2})), we further constrain our representation as:
\begin{equation}\label{eq:4}
\m E (\Delta) = \underset{\Delta}  {\text{minimize}} \sum_{(\m i, \m j)}^{\m K} \m w_{\m i\m j} \frac{1}{2}\| \Pi(\Theta_{\m i})-\Pi(\Theta_{\m j})\|_2^{\m F}
\end{equation}
The parameter `$\m w_{\m i \m j}$' accommodate the similarity knowledge between the two Grassmannians. Using the {\bf{Definition}}(\ref{def:2}) and Eq.(\ref{eq:3}), we further simplify Eq.(\ref{eq:4}) as   
\begin{equation}\label{eq:5}
\begin{aligned}
& \displaystyle \m E (\Delta) \equiv \underset{\Delta}  {\text{minimize}} \sum_{(\m i, \m j)}^{\m K} \m w_{\m i\m j}\frac{1}{2}\|\Delta^{\m T} \Omega_{\m i} \Omega_{\m i}^{\m T} \Delta -\Delta^{\m T} \Omega_{\m j} \Omega_{\m j}^{\m T} \Delta \|_{\m F}^{2} \\
& \displaystyle \m E (\Delta) \equiv \underset{\Delta}  {\text{minimize}} \sum_{(\m i, \m j)}^{\m K} \m w_{\m i\m j}\frac{1}{2} \|\Delta^{\m T}(\Omega_{\m i} \Omega_{\m i}^{\m T} - \Omega_{\m j} \Omega_{\m j}^{\m T}) \Delta  \|_{\m F}^{2} \\
& \displaystyle \m E (\Delta) \equiv \underset{\Delta}  {\text{minimize}} \sum_{(\m i, \m j)}^{\m K} \m w_{\m i\m j}\frac{1}{2} \|\Delta^{\m T}(\Lambda_{\m i\m j})\Delta  \|_{\m F}^{2} \\
\end{aligned}
\end{equation}

where, $\Lambda_{\m i \m j} \in Sym({d})$. The parameter `$\m w_{\m i \m j}$' (similarity graph) is set as $exp(-d_{g}^{2}(\Phi_{\m i}, \Phi_{\m j}))$ with $d_{g}$ as the projection metric (see {\bf{Definition}} (\ref{def:2})). Eq.(\ref{eq:5}) is an unconstrained optimization problem and its solution may provide a trivial solution. To estimate the useful solution, we further constrain the problem. Using $\m i^{\m t \m h}$ Grassmann point `$\Omega_{\m i}$' and its neighbors, expand Eq.(\ref{eq:5}). By performing some simple algebraic manipulation, Eq.(\ref{eq:5}) reduces to 
\begin{equation}\label{eq:6}
\mathbf{trace}\Big(\Delta^{\m T}\big(\sum_{\m i = 1}^{\m K}\lambda_{\m i \m i}\Omega_{\m i}\Omega_{\m i}^{\m T}\big)\Delta\Big)
\end{equation}
where, $\lambda_{\m i\m i} = \sum_{\m j = 1}^{\m K} \m w_{\m i \m j}$. Constraining the value of Eq.($\ref{eq:6}$) to 1 provides the overall optimization for an efficient representation of the local non-rigid surface on the Grassmann manifold.
\begin{equation}\label{eq:7}
\begin{aligned}
& \displaystyle \m E(\Delta) \equiv \underset{\Delta}  {\text{minimize}} \sum_{(\m i, \m j)}^{\m K} \m w_{\m i\m j}\frac{1}{2} \|\Delta^{\m T}(\Lambda_{\m i\m j})\Delta  \|_{\m F}^{2}\\
& \displaystyle \textrm{subject to:}\\
& \displaystyle \mathbf{trace}\Big(\Delta^{\m T}\big(\sum_{\m i = 1}^{\m K}\lambda_{\m i \m i}\Omega_{\m i}\Omega_{\m i}^{\m T}\big)\Delta\Big) = 1
\end{aligned}
\end{equation}
Its easy to verify that the matrix $\Lambda$ and $\big(\sum_{\m i = 1}^{\m K}\lambda_{\m i \m i}\Omega_{\m i}\Omega_{\m i}^{\m T}\big)$ are symmetric and positive semi-definite, and therefore, the above optimization can be solved as a generalized eigen value problem ---refer supplementary material for details.
\subsection{Dense NRSfM formulation}
To solve the dense non-rigid structure from motion with the representation formulated in the previous sub-section \S \ref{ss:gr_rep}, we propose to jointly optimize the objective function over the 3D structure and its local group representation. In order to build the overall objective function, we introduce each constraint equation one by one for clear understanding of our overall cost function.
\begin{equation}\label{eq:8}
\mathbf{E}_{\m p}(\m S) = \underset{\m S}  {\text{minimize}} ~\frac{1}{2}\|\m W - \m R \m S\|_{\m F}^2
\end{equation}
The {\bf{first}} term constrain the 3D structure such that it satisfies the re-projection error. 

\begin{equation}\label{eq:9}
\begin{aligned}
& \displaystyle \mathbf{E}_{\m s}(\m S^{\sharp}) = \underset{{\m S^{\sharp}}}  {\text{minimize}} ~\| \m S^{\sharp}\|_*
\end{aligned}
\end{equation}

The {\bf{second}} term caters the global assumption about the non-rigid object; that is the overall shape matrix is low-rank. To establish this assumption, we perform rank minimization of the shape matrix. Although the rank minimization of a matrix is NP-hard, it's relaxed to nuclear norm minimization to find an approximate solution. This term mainly penalizes the total number of independent shape required to represent the shape. The choice of minimizing $\m S^{\sharp} \in \mathbb{R}^{3 \m P \times \m F}$  instead to $\m S \in \mathbb{R}^{3 \m F \times \m P}$ is inspired from Dai \etal's work \cite{dai2014simple}. Since the dense deforming shape is composed of several local linear low-dimensional subspace, the global constraint (Eq.(\ref{eq:9})) may not reflect their local dependency. Therefore, in order to introduce the local subspace constraint on the shape, we use the notion of self-expressiveness on the non-linear Grassmann manifold space.
\begin{equation}\label{eq:10}
\begin{aligned}
& \displaystyle \underset{\m E, {\m C}, \m S^{\sharp}}  {\text{minimize}} ~\| \m E \|_{\mathcal{G}}^2 + \beta_2\| \m S^{\sharp}\|_* + \beta_3 \| \m C\|_*\\
& \displaystyle \textrm{subject to:} ~\m S^{\sharp} = f(\m S), \m S = \m S \m C + \m E
\end{aligned}
\end{equation}

Here, we define $f: \m S \in \mathbb{R}^{3 \m F \times \m P} \mapsto \m S^{\sharp} \in \mathbb{R}^{3 \m P \times \m F}$ and $\m C \in \mathbb{R}^{\m P \times \m P}$ as the coefficient matrix. We know from the literature that the Grassmann manifold is isometrically equivalent to the symmetric idempotent matrix \cite{chikuse2012statistics}. So, we embed the Grassmann manifold into symmetric matrix manifold to define the self-expressiveness. Let $\tilde{\xi} = \{ \Theta_1, \Theta_2, ..., \Theta_{\m K} \}$ be the set of Grassmannians on a low-dimensional Grassmann manifold. The elements of $\tilde{\xi}$ are the projection of high-dimensional Grassmannian representation of the columns of `$\m S$' matrix. Let $\rchi = \{ (\Theta_1\Theta_1^{\m T}), (\Theta_2\Theta_2^{\m T}), ..., (\Theta_{\m K}\Theta_{\m K}^{\m T}) \}$ be its embedding onto symmetric matrix manifold. Using such embedding techniques we re-write Eq.(\ref{eq:10}) as

\begin{equation}\label{eq:11}
\begin{aligned}
& \displaystyle \underset{\m E, \tilde{\m C}, \m S^{\sharp}}  {\text{minimize}} ~ \| \m E \|_{\m F}^2 + \beta_2\| \m S^{\sharp}\|_* + \beta_3 \| \tilde{\m C} \|_* \\
& \displaystyle \textrm{subject to:} \m S^{\sharp} = f(\m S), \rchi = \rchi \tilde{\m C} + \m E
\end{aligned}
\end{equation}
where, $\tilde{\m C} \in \mathbb{R}^{\m K \times \m K}$  and $\rchi \in \mathbb{R}^{\tilde{\m d} \times \tilde{\m d} \times \m K}$ denotes the coefficient matrix of Grassmannians and structure tensor respectively, with $\m K$ as the total number of Grassmannians. Generally, $\m K<<\m P$, which makes such representation scalable.

The {\bf{third}} term we introduce is composed of few  constraint functions that provides a way to group Grassmannians and recover 3D shape simultaneously. Let $\mathbf{P}$ $\in \mathbb{R}^{1 \times \m P}$ be an ordering vector that contains the index of columns of $\m S$. Our function definition is of the form $\{(\textrm{output}, \textrm{function(.)}): \textrm{definition} \}$. Using it, we define the function $f_g$, $f_h$, $f_{p}$ and $f_s$ as follows:
\begin{equation}\label{eq:12}
\begin{aligned}
& \displaystyle \big\{ \big({\xi}, f_g(\mathbf{P}, \m S)\big): \textrm{order} ~\{\m S_{\m i}\}_{\m i=1}^{\m K} ~\text{columns of} ~\m S ~\text{of using} ~{\mathbf{P} }, \\
& \displaystyle ~\xi := \{\Phi_{\m i} \}_{\m i = 1}^{\m K} \text{where}, [~\Phi_{\m i}, \Sigma_{\m i}, \xi_\textrm{vi} ] = \textrm{svds}(\m S_{\m i}, \m p) \big\}
\end{aligned}
\end{equation}
\begin{equation}\label{eq:13}
\begin{aligned}
& \displaystyle \big\{ \big(\tilde{\xi}, f_h(\Delta, \xi)\big): \tilde{\xi} = \{\Theta_{\m i}\}_{\m i=1}^{\m K}, \Theta_{\m i} = \Delta^{\m T}(\Phi_{\m i} \m U_{\m i}^{-1}), \\ 
& \displaystyle \textrm{where,} ~\Delta  = \text{solution to the minimization of Eq.}(\ref{eq:7})\big\}
\end{aligned}
\end{equation}
\begin{equation}\label{eq:14}
\begin{aligned}
& \displaystyle \big\{(\mathbf{P}, f_{p}(\tilde{\xi}, \tilde{\m C}, \mathbf{P}_{\m o}): \mathbf{P} = \text{spectral\_clustering}(\tilde{\xi}, \tilde{\m C}, \mathbf{P}_{\m o})\big\} 
\end{aligned}
\end{equation}
\begin{equation}\label{eq:15}
\begin{aligned}
& \displaystyle \big\{\big(\m S, f_s({\xi}, \Sigma, {\xi}_{\m v})\big): \m S_{\m i} = [{\xi}_{\m i}~{\Sigma_{\m i}}~{\xi}_{\m v\m i}], \text{where} ~{\Sigma_{\m i}} \in \mathbb{R}^{\m p \times \m p} \big\}
\end{aligned}
\end{equation}
Intuitively, the first function ($f_g$) uses the ordering vector $\mathbf{P}$ $\in \mathbb{R}^{1 \times \m P}$ to refine the grouping of the trajectories for suitable Grassmannian representation. The second function ($f_h$) projects the Grassmannians to a lower dimension in accordance with the neighbors using Eq.(\ref{eq:7}). The third function ($f_{p}$) uses the projected Grassmannians to assign proper labeling to the Grassmann points and update the given ordering vector $\mathbf{P}$ using spectral clustering. The fourth function ($f_s$) uses the group of trajectories to reconstruct back the set of local surface. ${\Sigma}$, ${\xi_{\m v}}$ are the singular values and right singular vector matrices in the high-dimension.
\\
\noindent
{\bf{Objective Function:}}
Combining all the above terms and constraints provides our overall cost function. 
\begin{equation}\label{eq:15}
\begin{aligned}
& \displaystyle \underset{\m E, \tilde{\m C}, \m S, \m S^{\sharp}}  {\text{minimize}} \frac{1}{2}\|\m W - \m R \m S\|_{\m F}^2 +  \beta_1\| \m E \|_{\m F}^2 + \beta_2\| \m S^{\sharp}\|_* + \beta_3 \| \tilde{\m C} \|_* \\
& \displaystyle \textrm{subject to:}\\
& \displaystyle \m S^{\sharp} = f(\m S), \rchi = \rchi \tilde{\m C} + \m E, \\
& \displaystyle {\xi} =  f_g(\mathbf{P}, \m S), \tilde{\xi} =  f_h(\Delta, \xi),  \\
& \displaystyle \m S =  f_s({\xi}, \Sigma, {\xi}_{\m v}), \mathbf{P} = f_p(\tilde{\xi}, \tilde{\m C}, \mathbf{P}_{\m o})
\end{aligned}
\end{equation}

where $\mathbf{P}_{\m o}$ vector contains the initial ordering of the columns of `$\m W$' and `$\m S$'. The function $(f_p)$ provides the ordering index to rearrange the columns of `$\m S$' matrix to be consistent with `$\m W$' matrix. This is important because, grouping the set of columns of `$\m S$' over iteration, disturbs its initial arrangements.

\section{Solution}
The optimization proposed in Eq.(\ref{eq:15}) is a coupled optimization problem. Several methods of Bi-level optimization can be used to solve such minimization problem \cite{bard2013practical, gould2016differentiating}. Nevertheless, we propose ADMM \cite{boyd2011distributed} based solution due to its application in many non-convex optimization problems. The key point to note is that one of our constraint is composed of separate optimization problem $(f_h)$ \ie, the solution to Eq.(\ref{eq:7}), and therefore, we cannot directly embed the constraint to the main objective function. Instead, we only introduce two Lagrange multiplier $\mathbf{L_1, L_2}$ to concatenate a couple of constraints back to the original objective function. The remaining constraints are enforced over iteration.  To decouple the variable $\tilde{\m C}$ from $\rchi$, we introduce auxiliary variable $\tilde{\m C} = \m Z$. We apply these operations to our optimization problem to get the following Augmented Lagrangian form:
\begin{equation}\label{eq:16}
\begin{aligned}
& \displaystyle \underset{\m Z, \tilde{\m C}, \m S, \m S^{\sharp}}  {\text{minimize}} \frac{1}{2}\|\m W - \m R \m S\|_{\m F}^2 +  \beta_1\| \rchi - \rchi \tilde{\m C} \|_{\m F}^2 + \beta_2\| \m S^{\sharp}\|_* + \\
&\displaystyle ~~~~~~~~~~~~~\frac{\rho}{2}\| \m S^{\sharp} - f(\m S)\|_{\m F}^2 + <\mathbf{L_1}, \m S^{\sharp} - f(\m S)> + \beta_3 \| \m Z \|_* + \\
&\displaystyle ~~~~~~~~~~~~~\frac{\rho}{2}\| \tilde{\m C} - \m Z\|_{\m F}^2 + <\mathbf{L_2}, \tilde{\m C} - \m Z>\\
& \displaystyle \textrm{subject to:} {~\xi} =  f_g(\mathbf{P}, \m S), \tilde{\xi} =  f_h(\Delta, \xi),  \\
& \displaystyle \m S =  f_s({\xi}, \Sigma, {\xi}_{\m v}), \mathbf{P} = f_p(\tilde{\xi}, \tilde{\m C}, \mathbf{P}_{\m o})
\end{aligned}
\end{equation}

\begin{algorithm}[t!]
\caption{\small {Dense Non-rigid Structure from Motion ({\bf{M}}\textcolor{red}{o}{\bf{J}}\textcolor{red}{u})}}
\label{algo:Algorithm1}
\begin{algorithmic}
\REQUIRE
$\m W$, $\m R$, $\{\beta_{\m i}\}_{\m i=1}^3$, $\rho$=$e^{-2}$, $\rho_m$=$e^{8}$, $\epsilon$=$e^{-10}$, $c$ =$1.1$, $\m K$; \\ 
\hspace{-0.3cm}{\bf Initialize:} ${\m S}$=$\mathbf{pinv}(\m R) \m W$, ${\m S^{\sharp}}$=$f(\m S)$, $\m Z$=$\mathbf{0}$, $\{{\bf L}_{\m i}\}_{\m i=1}^2$=$\mathbf{0}$, $\tilde{\m d}$;\\ 
\hspace{0.5cm}$\Delta$ = $[\mathbf{I}_{\tilde{\m d} \times \tilde{\m d}};$ $\textrm{random values}],$ $\m p $ $\%$top singular values \\
\hspace{0.5cm}$\mathbf{P}_{\m o}$ = $\textrm{kmeans++}(\m S, \m K)$, $\textrm{iter}$ = 1, $\mathbf{P}_\textrm{store}(\textrm{iter}, :)$ = $\mathbf{P}_{\m o}$,\\
\hspace{0.5cm}$\mathbf{P}$ = $\mathbf{P}_{\m o}$
\WHILE {not converged}
{\small
\STATE 1. $\m S$ := mldivide\big($\m R^{\m T}\m R + \rho \m I$, $ \rho(f^{-1}(\m S^{\sharp}) + \frac{f^{-1}(\mathbf{L_1})}{\rho}) + \m R^{\m T}\m W$\big);
\STATE 2. $\xi := f_g(\mathbf{P}, \m S)$; see Eq.(\ref{eq:12})
\STATE 3. $\m W := \text{arrange\_column}(\mathbf{P}, \m W)$
\STATE 4. Update the similarity matrix `$\m w_\textrm{ij}$' using $\xi$. \S \ref{ss:gr_rep}
\STATE 5. $\tilde{\xi} := f_{h}(\xi, \Delta); \text{s.t}, \Delta \equiv \underset{\Delta}{\textrm{minimize}}~~\m E(\Delta);$  see Eq.(\ref{eq:13})
\STATE 6. $\Gamma_{\m i  \m j}=\textrm{Tr}[(\Theta_{\m j}^{\m T}\Theta_{\m i})((\Theta_{\m i}^{\m T}\Theta_{\m j})];\Gamma=(\Gamma_{\m i \m j})_{\m i\m j=1}^{\m K};\m L = \textrm{chol}(\Gamma)$ \vspace{-0.2cm}
\STATE 7. $ \tilde{\m C}$ := $\big(2\beta_1 \m L \m L^{\m T} + \rho(\m Z - \frac{\mathbf{L_2}}{\rho})\big)$ $\big(2\beta_1 \m L \m L^{\m T} + \rho \m I\big)^{-1}$;
\STATE 8. $\mathbf{P} := f_{p}(\tilde{\xi}, \tilde{\m C}, \mathbf{P})$;

\STATE 9. $\m S := f_s({\xi}, {\Sigma}, {\xi}_v);$  see Eq.(\ref{eq:14})
\STATE 10. $\m S^{\sharp}:= \m U_{\m s}\mathcal{S}_{\frac{\beta_2}{\rho}}(\Sigma_{\m s})\m V_{\m s}; \text{s.t},[\m U_{\m s},\Sigma_{\m s},\m V_{\m s}] := \textrm{svd}(f(\m S) - \frac{\mathbf{L_1}}{\rho})$\vspace{-0.2cm}
\STATE 11. $\m Z := \m U_{\m z}\mathcal{S}_{\frac{\beta_3}{\rho}}(\Sigma_{\m z})\m V_{\m z}; \text{s.t}, [\m U_{\m z},\Sigma_{\m z},\m V_{\m z}] := \textrm{svd}(\tilde{\m C} + \frac{\mathbf{L_2}}{\rho});$
\STATE 12. $\mathbf{L_1} := \mathbf{L_1} + \rho(\m S^{\sharp} - f(\m S)); \mathbf{L_2} := \mathbf{L_2} + \rho(\tilde{\m C} - \m Z)$
\STATE 13. $\textrm{iter} := \textrm{iter} + 1;$ $\mathbf{P}_\textrm{store}(\textrm{iter}, :) := \mathbf{P};$ 
\STATE 14. $\rho := \textrm{min}(\rho_m, c\rho);$ \\
\STATE 15. $\textrm{gap} := ~~\textrm{max}\{\|\m S^{\sharp} - f(\m S)\|_\infty, \| \tilde{\m C} - \m Z\|_\infty\};$\\
$(\textrm{gap}<\epsilon) \lor (\rho > \rho_m)\rightarrow \textbf{break};\textrm{\%convergence check}$
} \\
\ENDWHILE
\RETURN $\m S$;\\
\hspace{-0.3cm}$e_{3 \m D}$ = \textbf{Estimate\_error} $(\m S, \m S_{\m G \m T}, \mathbf{P}_\textrm{store});$  \%use Eq.(\ref{eq:17})
\end{algorithmic}
\end{algorithm}
Note that $\tilde{\m C}$ provides the information about the subspace, not the vectorial points. However, we have the chart of the trajectories and its corresponding subspace. Once, we group the trajectories based on $\tilde{\m C}$, $f_g(.)$ provides new Grassmann sample corresponding to each group. The definition of $f_h(.)$ and $f_s(.)$ is provided in Eq.(\ref{eq:7}) and Eq.(\ref{eq:14}) respectively. More generally, the solution to the optimization in Eq.(\ref{eq:7}) is obtained by solving it as a generalized eigenvalue problem. To keep the order of columns of `$\m S$' matrix consistent with `$\m W$' matrix $f_p(.)$ provides the ordering index. We provide the implementation details of our method with suitable MATLAB commands in the Algorithm Table (\ref{algo:Algorithm1}). For details on the derivation to each sub-problem, kindly refer to the supplementary material.

\section{Initialization and Evaluation}
We performed experiments and evaluation on the available standard benchmark datasets \cite{garg2013dense,varol2012constrained,beeler2011high}. To keep our evaluations consistent with the previous methods, we compute the mean normalized 3D reconstruction error of the estimated shape `$\m S_\textrm{est}$' after convergence as 
\begin{equation}\label{eq:17}
\m e_{3 \m D} = \frac{1}{\m F} \sum_{\m i = 1}^{\m F}\frac{\|\m S_\textrm{ est}^{\m i} - \m S_{\m G\m T}^{\m i}\|_{\m F}}{\|\m S_{\m G\m T}^{\m i}\|_{\m F}}
\end{equation}
here `$\m S_{\m G\m T}$' denotes the ground-truth 3D shape matrix.
\paragraph{Initialization:} We used Intersection method \cite{dai2014simple} to estimate the rotation matrix and initialize $\m S = \textbf{pinv}(\m R)\m W$. The initial grouping of the trajectories or columns of $\m S$ is done using k-means++ algorithm \cite{arthur2007k}. These groups are then used to initialize $\mathbf{P}_{\m o}$, $\mathbf{P}$ and the Grassmann points $\{\Phi_{\m i} \}_{\m i=1}^{\m K} \in \xi$ via subset of singular vectors. To represent the Grassmannians in the lower-dimension, we solve Eq.(\ref{eq:7}) to initialize $\tilde{\xi}$ and store corresponding singular values. The similarity matrix or graph in Eq.(\ref{eq:7}) is build using the distance measure between the Grassmannians in the embedding space \S \ref{ss:gr_rep}.\\
\noindent
{\bf{1. Results on synthetic Face dataset:}}
The synthetic face dataset is composed of four distinct sequence \cite{garg2013dense} with 28,880 feature points tracked over multiple frames. Each sequence captures the human facial expression with a different range of deformations and camera motion. Sequence 1 and Sequence 2 are 10 frame long video with rotation in the range $\pm 30^{\circ}$ and $\pm 90^{\circ}$ respectively. Sequence 3 and Sequence 4 are 99 frame long video that contains high frequencies and low frequencies rotation respectively which captures real human facial deformations. Table (\ref{tab:statisticalresults}) shows the statistical results obtained on these sequences using our algorithm. For qualitative results on these sequences kindly refer to the supplementary material.

\begin{figure}[t]
\begin{center}
\includegraphics[width=1.0\linewidth, height= 0.60\linewidth]{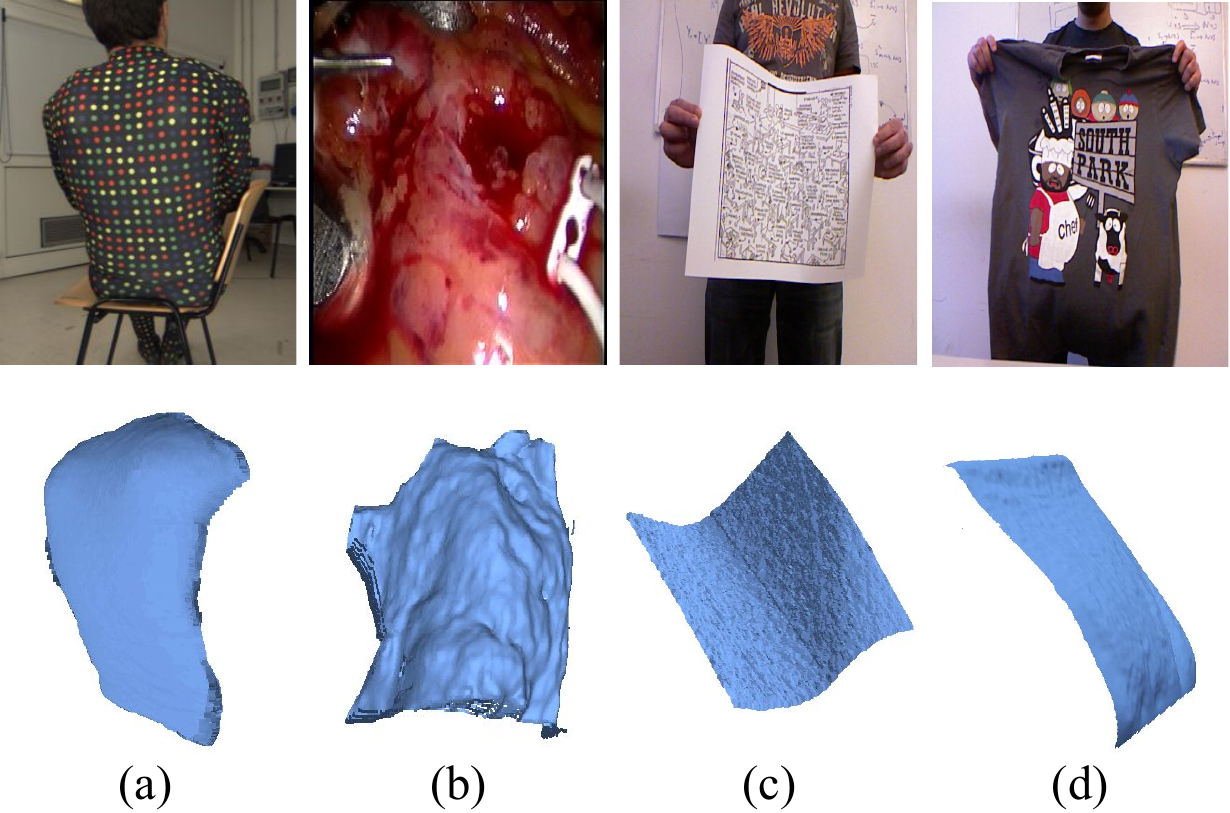}
\caption{\small{ From left: 3D reconstruction results on Back \cite{garg2013dense}, Heart \cite{garg2013dense}, Paper\cite{varol2012constrained} and T-shirt \cite{varol2012constrained} data sequence respectively.} \label{fig:multidataset}}
\end{center}
\end{figure}
\noindent
{\bf{2. Results on Paper and T-shirt dataset:}}
Varol \etal introduced `kinect\_paper' and `kinect\_tshirt' datasets to test the performance of NRSfM algorithm under real conditions \cite{varol2012constrained}. This dataset provides sparse SIFT \cite{lowe1999object} feature tracks and noisy depth information captured from Microsoft Kinect for all the frames. As a result, to get dense 2D feature correspondences of the non-rigid object for all the frames becomes difficult. To circumvent this issue, we used Garg \etal \cite{garg2011robust} algorithm to estimate the measurement matrix. To keep the numerical comparison consistent with the previous work in dense NRSfM \cite{kumar2018scalable}, we used the same coordinate range for tracking the features. Numerically, its $\m x_{\m w}$ = $(253, 253, 508, 508)$, $\m y_{\m w}$ = $(132, 363, 363, 132)$ rectangular window across 193 frames for kinect\_paper sequence. For kinect\_tshirt sequence, we considered rectangular window of $\m x_{\m w}$ = $(203, 203, 468, 468)$, $\m y_{\m w}$ = $ (112, 403, 403, 112)$ across 313 frames, same as used in Kumar \etal work \cite{kumar2018scalable}. Fig.(\ref{fig:multidataset}) shows the reconstruction results on these sequence with comparative results provided in Table (\ref{tab:statisticalresults}).

\begin{figure}[t]
\begin{center}
\includegraphics[width=1.0\linewidth, height = 0.60\linewidth]{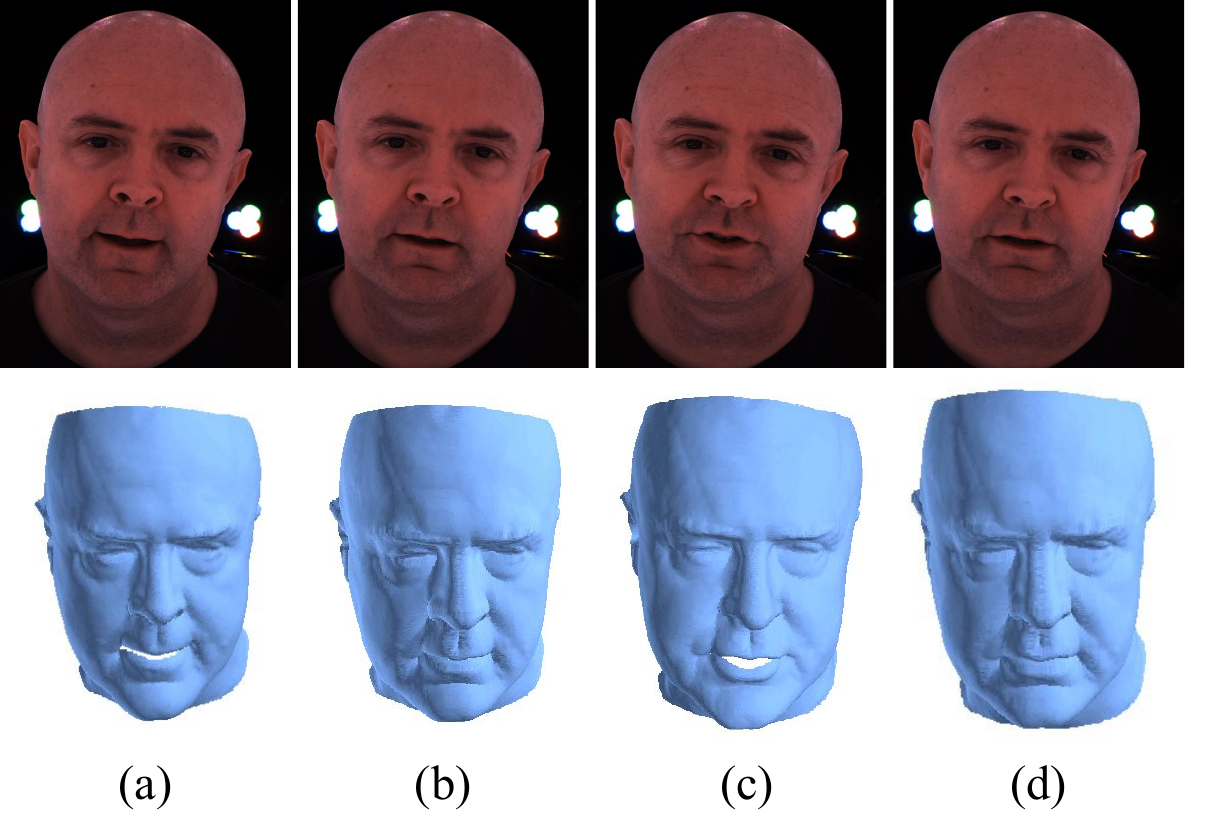}
\caption{\small{ 3D reconstruction results on the Actor sequence \cite{beeler2011high}.} \label{fig:Actor2}}
\end{center}
\end{figure}

\begin{table*}[h]
\centering
\small
\begin{tabular}{|>{\columncolor[gray]{0.88}}c|c|c|c|c|c|c|c|c|c|}
\hline
\rowcolor[gray]{0.75}
Dataset $\downarrow \slash$ Method $\rightarrow$      & MP \cite{paladini2012optimal}  & PTA \cite{akhter2009nonrigid}  & CSF1 \cite{gotardo2011computing} & CSF2\cite{gotardo2011non} &DV \cite{garg2013dense} & DS \cite{dai2017dense} & SMSR \cite{ansari2017scalable}  & SDG\cite{kumar2018scalable} & Ours \\ \hline
Face Sequence 1 &   0.2572  &   0.1559  & 0.5325    & 0.4677 & 0.0531 & 0.0636 & 0.1893 & 0.0443 & {\bf{0.0404}}\\ \hline
Face Sequence 2 &   0.0644  &   0.1503  & 0.9266    & 0.7909 & 0.0457 & 0.0569  & 0.2133  & {\bf{0.0381}} & 0.0392 \\ \hline
Face Sequence 3 &   0.0682 &   0.1252  & 0.5274    & 0.5474 & 0.0346 & 0.0374 & 0.1345  &  0.0294 & {\bf{0.0280}}\\ \hline
Face Sequence 4 &   0.0762 &   0.1348  & 0.5392    & 0.5292 & 0.0379 &  0.0428 & 0.0984 &  {\bf{0.0309}}  & 0.0327\\ \hline
Actor Sequence 1 &  0.5226  &   0.0418  & 0.3711    & 0.3708 & - & 0.0891 & 0.0352 & 0.0340 & {\bf{0.0274}} \\ \hline
Actor Sequence 2 &  0.2737  &   0.0532  & 0.2275    & 0.2279 & - & 0.0822 & 0.0334 & 0.0342 & {\bf{0.0289}} \\  \hline
Paper Sequence   &   0.0827 &   0.0918  &  0.0842   & 0.0801 &  - & 0.0612 & - & 0.0394 & {\bf{0.0338}} \\ \hline
T-shirt Sequence  &  0.0741 &   0.0712  &  0.0644   & 0.0628 &  - & 0.0636 & - & {\bf{0.0362}} & 0.0386 \\ \hline
\end{tabular}
\caption{ \small{Statistical comparison of our method with other competing approaches. Quantitative evaluations for SMSR \cite{ansari2017scalable} and DV \cite{garg2013dense} are not performed by us due to the unavailability of their code, and therefore, we tabulated their reconstruction error from their published work.}} \label{tab:statisticalresults}
\end{table*}

\begin{figure*}
\centering
\subfigure [\label{fig:n1}] 
{\includegraphics[width=0.24\textwidth, height=0.13\textheight]{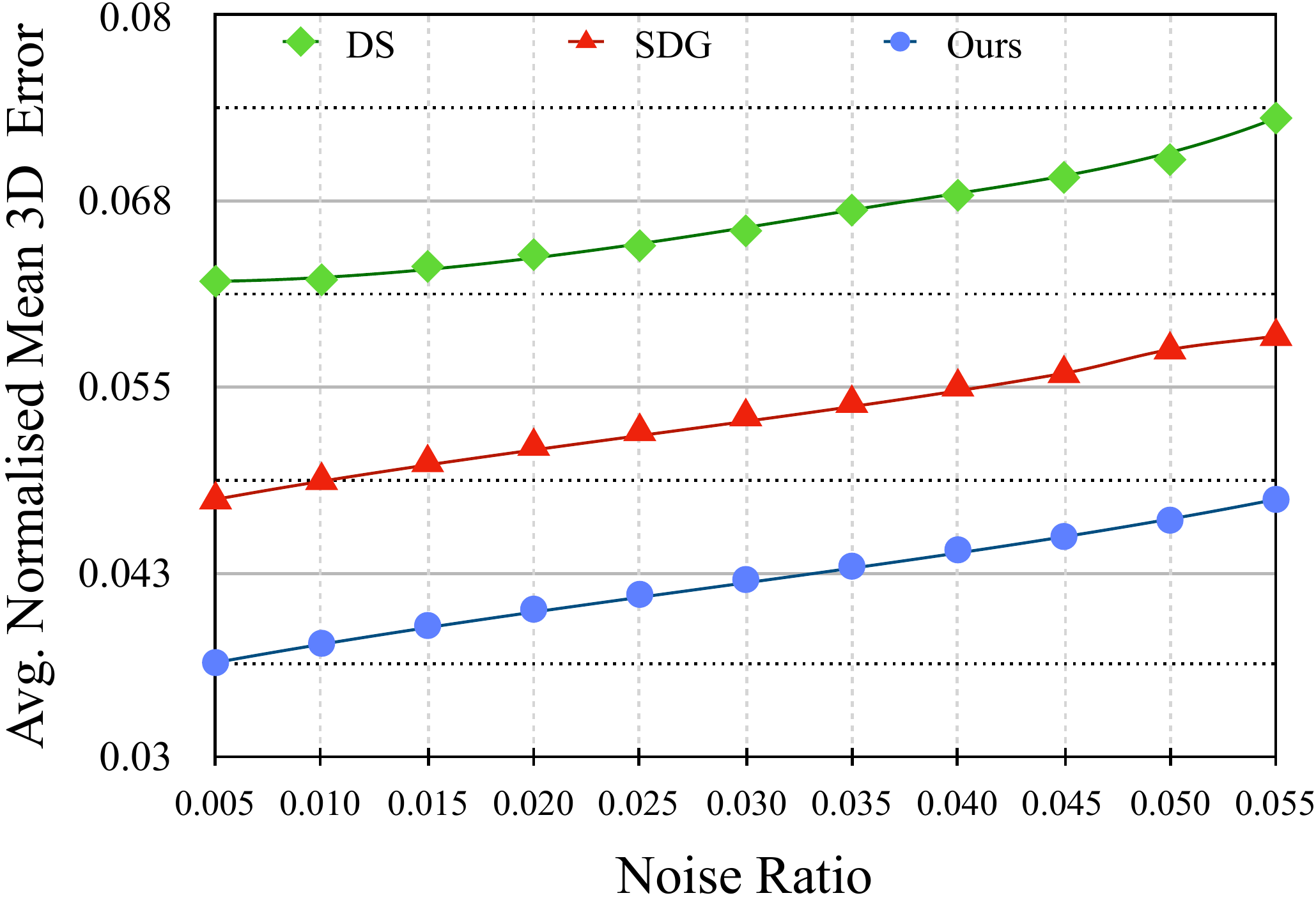}}
\subfigure [\label{fig:m1}] {\includegraphics[width=0.24\textwidth, height=0.13\textheight]{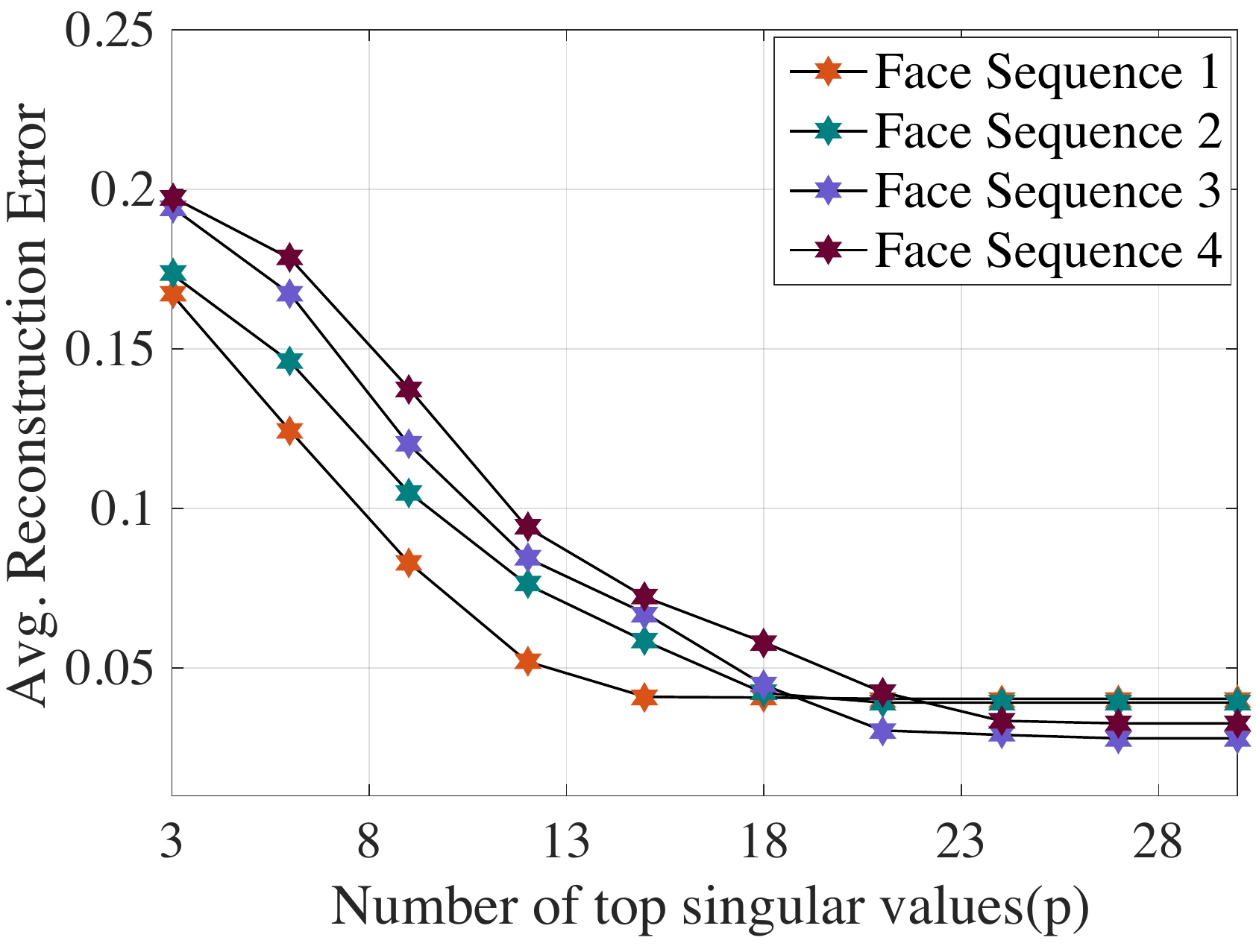}}
\subfigure [\label{fig:o1}] {\includegraphics[width=0.24\textwidth, height=0.13\textheight]{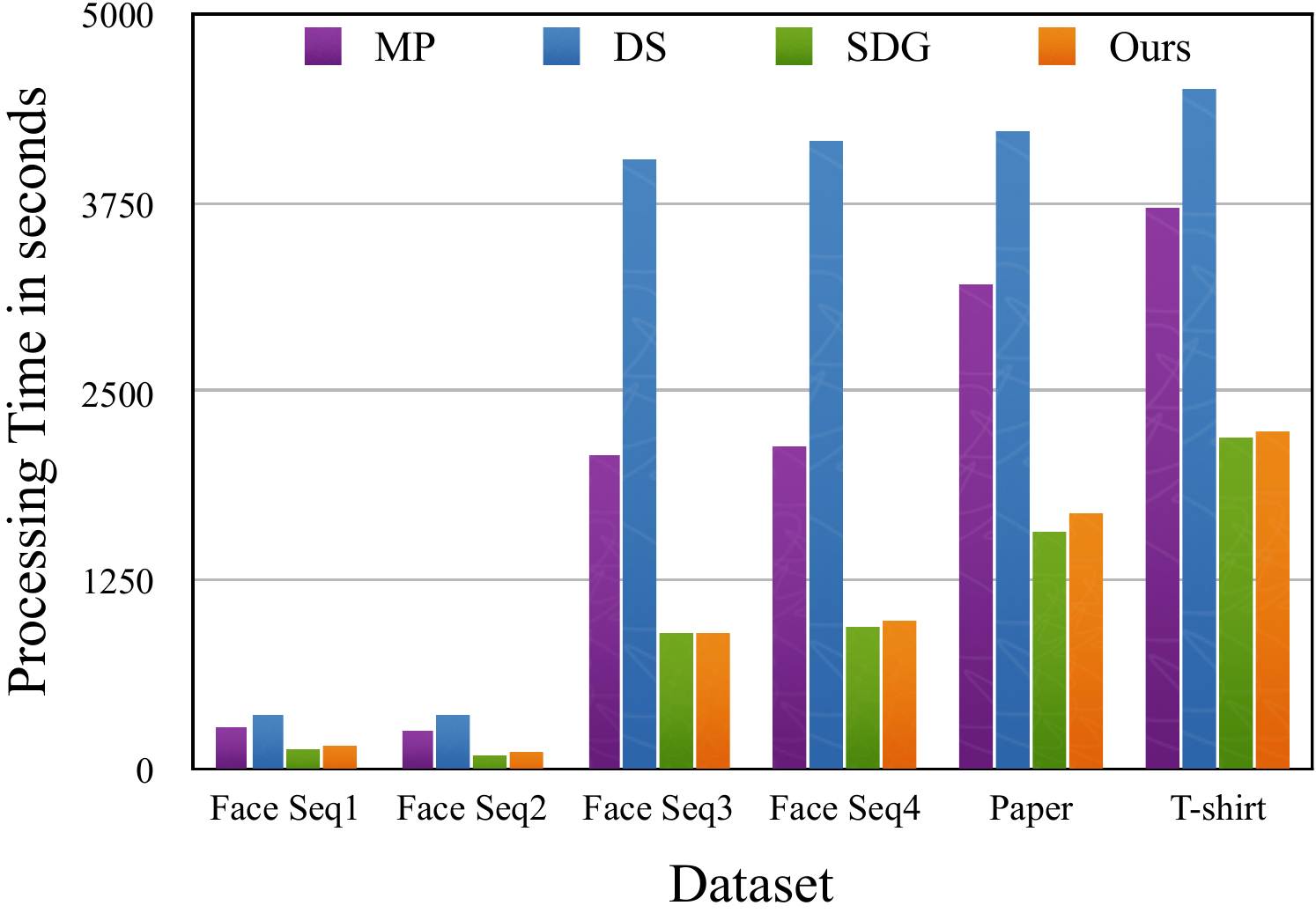}}
\subfigure [\label{fig:p1}] {\includegraphics[width=0.24\textwidth, height=0.13\textheight]{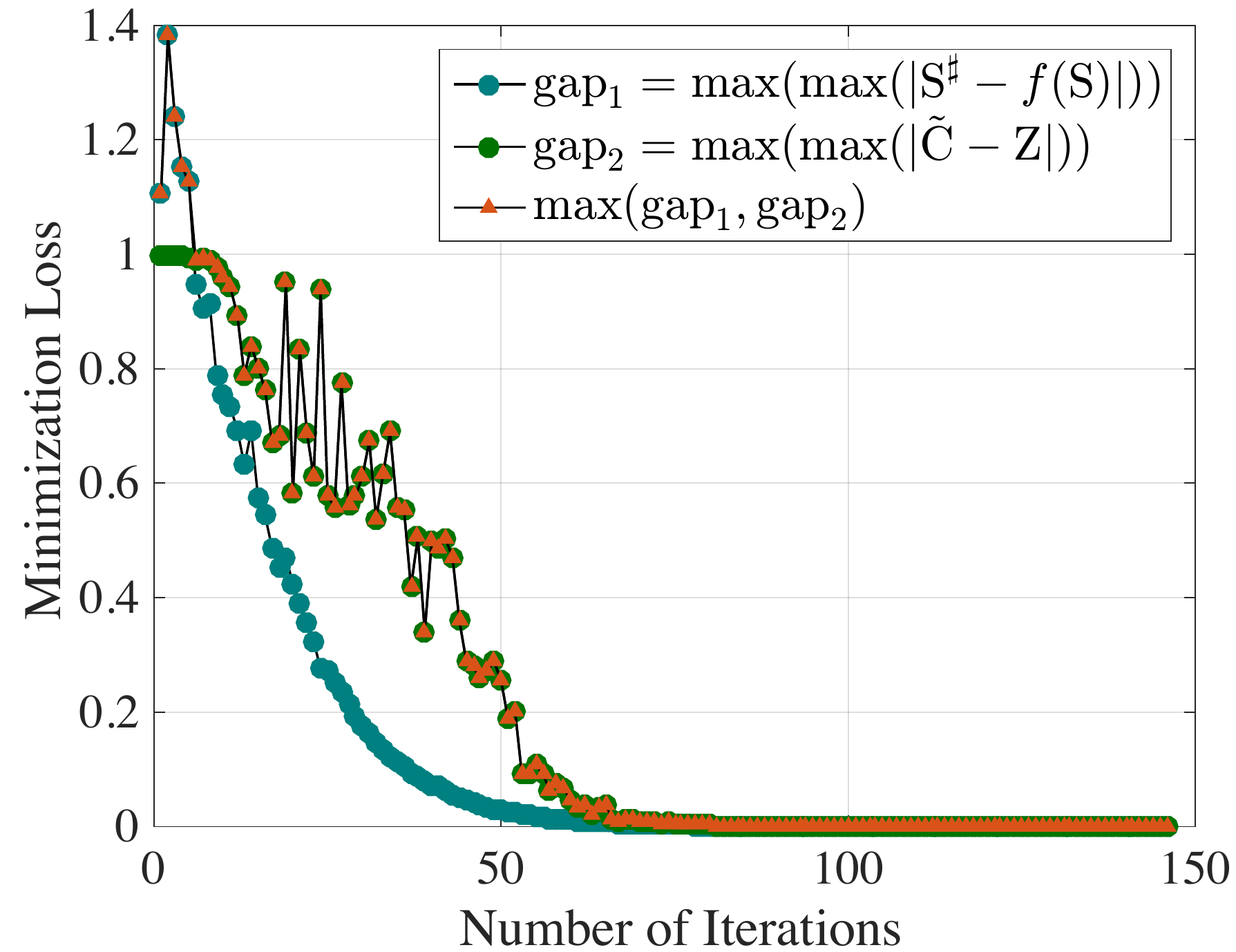}}
\caption{\small{(a) Variation in the average 3D reconstruction error with change in the noise ratio's for face dataset\cite{garg2013dense}. (b) Fluctuation in the 3D reconstruction accuracy with change number of top singular values and corresponding singular vectors used by our algorithm for face sequence\cite{garg2013dense}. (c) Processing time againt other competing algorithm's on Intel Core i7-4790 CPU @3.60GHz x 8 Desktop with MATLAB 2017b, our method show comparable execution timing to SDG\cite{kumar2018scalable}. (d) A typical ADMM optimization convergence curve of our algorithm. }}
\label{fig:evaluationNoiseandMD}
\end{figure*}
\noindent
{\bf{3. Results on Actor dataset:}}
Beeler \etal \cite{beeler2011high} introduced Actor dataset for high-quality facial performance capture. This dataset is composed of 346 frames captured from seven cameras with 1,180,232 vertices. The dataset captures the fine details of facial expressions which is extremely useful in the testing of NRSfM algorithms. Nevertheless, for our experiment, we require dense 2D image feature correspondences across all images as input, which we synthesized using ground-truth 3D points and synthetically generated orthographic camera rotations. To maintain the consistency with the previous works in dense NRSfM for performance evaluations, we synthesized two different datasets namely Actor Sequence1 and Actor Sequence2 based on the head movement as described in Ansari \etal work \cite{ansari2017scalable}. Fig.(\ref{fig:Actor2}) shows the dense detailed reconstruction that is achieved using our algorithm. Table (\ref{tab:statisticalresults}) clearly indicates the benefit of our approach to reconstruct such complex deformations.

\noindent
{\bf{4. Results on Face, Heart, Back dataset:}}
To evaluate the variational approach to dense NRSfM \cite{garg2013dense} Garg \etal introduced these datasets. Its sequences are composed of monocular video's captured in a natural environment with varying lighting condition and large displacements. It consists of three different videos with 120, 150 and 80 frames for face sequence, back sequence and heart sequence respectively. Additionally, it provides dense 2D feature track for the same with 28332, 20561, and 68295 features track over the frames for face, back and heart sequence. No ground-truth 3D is available with this dataset for evaluation. Fig.(\ref{fig:multidataset}) show reconstruction results on back and heart sequence. For more qualitative results on these sequences, kindly refer to the supplementary material.

\subsection{Algorithmic Analysis}
\noindent
We performed some more experiments to understand the behavior of our algorithm under different input parameters and evaluation setups. In practice these experiment help analyze the practical applicability of our algorithm.\\
\noindent
{\bf{1. Performance over noisy trajectories:}}
We utilized the standard experimental procedure to analyze the behavior of our algorithm under different noise levels. Similar to the work of Kumar \etal \cite{kumar2018scalable}, we added the Gaussian noise to the input trajectories. The standard deviation of the noise are adjusted as $\sigma_{g} = \lambda_g\textrm{max}\{|\m W|\}$ with $\lambda_g$ varying from 0.01 to 0.055. Fig.(\ref{fig:n1}) show the quantitative comparison of our approach with recent algorithms namely DS \cite{dai2017dense} and SDG \cite{kumar2018scalable}. The graph is plotted by taking the average reconstruction error of all the four synthetic face dataset \cite{garg2013dense}. The procured statistics indicate that our algorithm is more resilient to noise than other competing methods.\\
\noindent
{\bf{2. Performance with change in the number of singular values:}}
The selection of `$\m p$' in $\mathcal{G}(\m p, \m d)$ \ie the number of top singular vectors for Grassmannian representation and its corresponding singular values to perform reconstruction can directly affect the performance of our algorithm. However, it has been observed over several experiments that we need very few singular value and singular vectors to recover dense detailed 3D reconstruction of the deforming object. Fig.(\ref{fig:m1}) show the variation in average 3D reconstruction with the values of `$\m p$' for synthetic face dataset \cite{garg2013dense}.\\
\noindent
{\bf{3. Processing Time and Convergence:}} Our algorithm execution time is almost at par or a bit slower than SDG \cite{kumar2018scalable}. Fig.(\ref{fig:o1}) show the processing time taken by our method on different datasets. Fig.(\ref{fig:p1}) show a typical convergence curve of our algorithm. Ideally, it takes 120-150 iteration to provide an optimal solution to the problem.



\section{Conclusion}
\noindent
Our Grassmannian representation of a non-rigidly deforming surface exploits the advantage of Grassmannians of different dimensions to jointly estimate better grouping of subspaces and their corresponding 3D geometry. Our approach explicitly leverages the geometric structure of the non-rigidly moving object w.r.t its neighbors on manifold via similarity graph and, it's embedding in the lower dimension. We empirically demonstrated that our method is able to achieve 3D reconstruction accuracy which is better or as good as the state-of-the-art, with significant improvement in handling noisy trajectories.

\noindent\textbf{Acknowledgement.}{\footnotesize  ~The author was supported in part by the ARC Centre of Excellence for Robotic Vision (CE140100016), ARC Discovery project on 3D computer vision for geo-spatial localisation (DP190102261) and ARC DECRA project (DE140100180). The author thank his elder brother Aditya Kumar for his relentless support and advise. Thank you Roya Safaei for her constant help and constructive suggestions. Also, the author thank Dr. J. Varela and Dr. C. Punch, ACT for the medical treatment during the course of this work}.

{\small
\balance
\bibliographystyle{ieee_fullname}
\bibliography{egbib}
\nocite{kumar2019motion}
\nocite{kumar2019simple}
}

\end{document}